\documentclass{article}

\usepackage{soul_style}
\usepackage[export]{adjustbox}
\usepackage[utf8]{inputenc} %
\usepackage[T1]{fontenc}    %
\usepackage{newunicodechar}
\usepackage{hyperref}       %
\usepackage{xcolor}
\usepackage[most]{tcolorbox}
\definecolor{logocolor}{HTML}{18C3B8}

\usepackage{amsthm}
\newtheorem{assumption}{Assumption}

\usepackage[normalem]{ulem} %
\hypersetup{
    colorlinks=true,      %
    linkcolor=blue,      %
    urlcolor=logocolor,       %
    citecolor=logocolor,      %
    linkbordercolor=blue, %
    urlbordercolor=blue,
    citebordercolor=blue,
    pdfborderstyle={/S/U/W 1}, %
}

\usepackage{float}
\usepackage{url}            %
\usepackage{booktabs}       %
\usepackage{amsfonts}       %
\usepackage{nicefrac}       %
\usepackage{microtype}      %
\usepackage{lipsum}		%
\usepackage{graphicx}
\usepackage{natbib}
\usepackage{doi}
\usepackage{amsmath}
\usepackage{amssymb} %
\usepackage{xspace}
\usepackage{enumitem}
\usepackage{multirow}
\usepackage{subcaption} 
\usepackage{makecell}
\usepackage{hyperref, cleveref}
\usepackage{pifont}
\usepackage[inkscapelatex=false]{svg}
\usepackage{caption}
\usepackage{wrapfig}
\usepackage{algorithm}
\usepackage{algorithmic}
\usepackage{threeparttable}
\usepackage{overpic}

\setlist[itemize]{leftmargin=*}
\setlist[enumerate]{leftmargin=*}
\setlist[description]{leftmargin=*}

\newcommand{\name}{SoulX-LiveAct\xspace}

\definecolor{midnightgreen}{rgb}{0.0, 0.29, 0.33}

\title{\textnormal{\name: 
Towards Hour-Scale Real-Time Human Animation with Neighbor Forcing and ConvKV Memory
}}

\author{
\normalfont Dingcheng Zhen$^*$$^\dagger$\textsuperscript{1}, 
Xu Zheng$^*$\textsuperscript{2}, 
Ruixin Zhang$^*$\textsuperscript{3}, 
Zhiqi Jiang$^*$\textsuperscript{1}, 
Yichao Yan\textsuperscript{1},
Ming Tao\textsuperscript{1}, 
Shunshun Yin\textsuperscript{1}   \\
\textsuperscript{1}Soul AI Lab, China \quad \textsuperscript{2}HKUST(GZ) \quad \textsuperscript{3}Soochow University\\
$*$: Equal Contribution, ${\dagger}$:  Corresponding Author
}

\renewcommand{\headeright}{\raisebox{-0.2\height}{\includegraphics[height=1.8em]{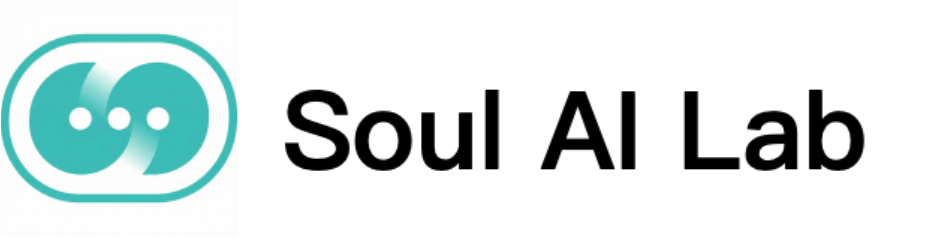}}}
\renewcommand{\shorttitle}{\name Technical Report}

\usepackage{amsmath}
\usepackage{CJKutf8}

\usepackage[most]{tcolorbox}
\usepackage{lipsum}

\newtcolorbox{abstractbox}{
  colback=logocolor!6,
  colframe=logocolor,
  boxrule=0.8pt,
  arc=6pt,
  left=12pt,
  right=12pt,
  top=10pt,
  bottom=10pt
}

\begin{document}
\maketitle
\vspace{-20pt}
\begin{abstractbox}
\begin{center}
\textbf{\large\bfseries Abstract}
\end{center}
Autoregressive (AR) diffusion models offer a promising framework for sequential generation such as video synthesis by combining diffusion modeling with causal inference. Although they support streaming generation, existing AR diffusion methods struggle to scale efficiently.
In this paper, we identify two key challenges in hour-scale real-time human animation. First, most forcing strategies propagate sample-level representations with mismatched diffusion states, causing inconsistent learning signals and unstable convergence. Second, historical representations grow unbounded and lack structure, preventing effective reuse of cached states and severely limiting inference efficiency.
To address these challenges, we propose \textbf{Neighbor Forcing}, a diffusion-step-consistent AR formulation that propagates temporally adjacent frames as latent neighbors under the same noise condition. This design provides a distribution-aligned and stable learning signal while preserving drifting throughout the AR chain. Building upon this, we introduce a structured \textbf{ConvKV memory} mechanism that compresses the keys and values in causal attention into a fixed-length representation, enabling constant-memory inference and truly infinite video generation without relying on short-term motion-frame memory. Extensive experiments demonstrate that our approach significantly improves training convergence, hour-scale generation quality, and inference efficiency compared to existing AR diffusion methods. Numerically, \name enables hour-scale real-time human animation and supports \textbf{\textit{20 FPS}} real-time streaming inference on as few as \textbf{two} NVIDIA H100 or H200 GPUs. Quantitative results demonstrate that our method attains state-of-the-art performance in lip-sync accuracy, human animation quality, and emotional expressiveness, with the lowest inference cost.

\vspace{8pt}

\textbf{Project Page:} \href{https://soul-ailab.github.io/soulx-liveact/}{https://soul-ailab.github.io/soulx-liveact/}

\textbf{Code:} \href{https://github.com/Soul-AILab/SoulX-LiveAct}{https://github.com/Soul-AILab/SoulX-LiveAct}
\end{abstractbox}

\section{Introduction}
Autoregressive (AR) diffusion models~\citep{henschel2024streamingt2v, liu2024mardini, InfinityStar, Yuan2025Lumos-1, zhen2025teller, zhen2025marrying} have emerged as a promising framework for sequential generation tasks such as video synthesis~\cite{xie2025progressive,yin2025slow,sun2025ar}. By combining diffusion modeling with causal AR generation, these methods support streaming inference and avoid the fixed-length constraints of full-sequence diffusion, making them suitable for scalable and online generation~\cite{huang2025selfforcing,liu2025rollingforcing}. Despite their success, existing AR diffusion methods differ substantially in how temporal information is propagated across time. Beyond whether a model is AR, a central question lies in \emph{what representation is propagated along the AR chain} and how it interacts with the diffusion process, as in Table~\ref{tab:ar_formulation_comparison}. This design choice critically affects both training dynamics and inference efficiency.

\begin{figure*}[t!]
     \centering
     \vspace{-10pt}
     \includegraphics[width=0.98\linewidth]{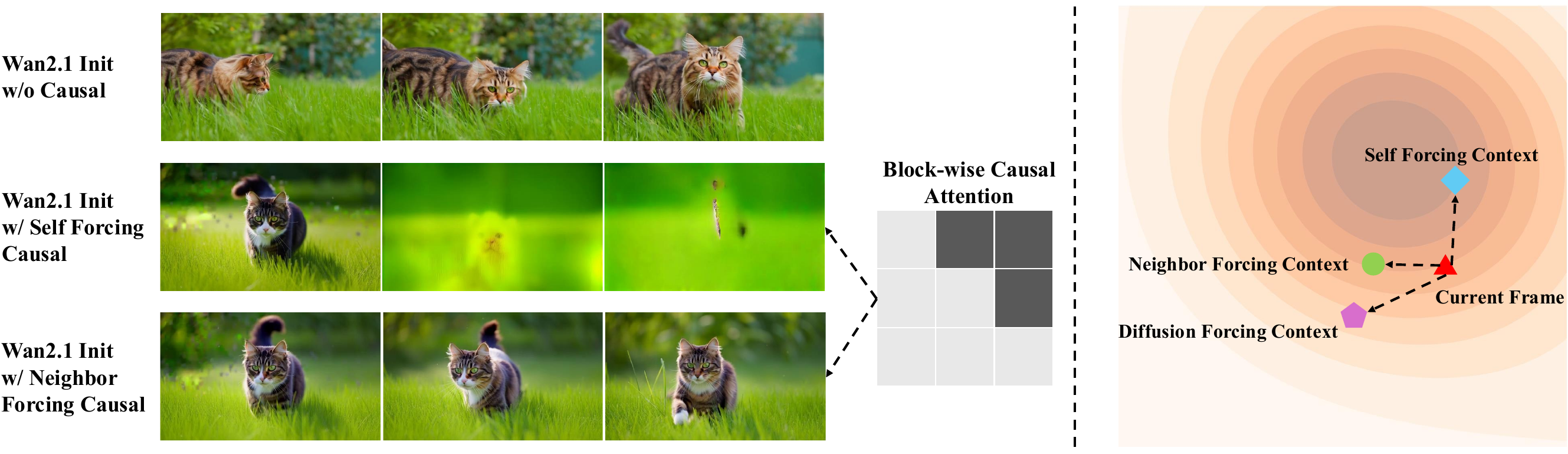}
    \caption{This phenomenon is observed when a causal attention mask is directly applied to a pretrained non-AR diffusion model, indicating that Neighbor Forcing effectively mitigates the mismatch between AR and non-AR diffusion.}
     \vspace{-15pt}
    \label{fig: intro} 
\end{figure*}

In our initial exploration, we observe an intriguing phenomenon shown in Figure~\ref{fig: intro}. When a \textbf{\textit{causal attention mask}} is directly imposed on a \textbf{\textit{pretrained non-AR diffusion model}}, existing forcing strategies—such as Diffusion Forcing~\cite{chen2024diffusionforcing} or Self Forcing~\cite{huang2025selfforcing}—fail to produce temporally coherent videos without additional training. In contrast, when the reference latent is chosen as the latent of the previous chunk evaluated at the \textbf{\textit{same diffusion step}}, the same non-AR model is able to generate subject-consistent and temporally stable videos in a zero-shot manner. Notably, this holds even though the original non-AR model exhibits noticeable chunk-level jitter.

\begin{table}[h!]
\centering
\caption{Comparison of AR diffusion-based methods from an AR formulation perspective.
\textbf{ARPP} denotes the representation propagated along the AR chain,
and \textbf{KV Reuse} indicates whether cached key--value states can be reused during inference.
Here, subscript $t$ denotes the diffusion step and superscript $f$ denotes the $f$-th frame latent.}
\setlength{\tabcolsep}{14pt}
\vspace{6pt}
\label{tab:ar_formulation_comparison}
\resizebox{\linewidth}{!}{
\begin{tabular}{l c c c}
\toprule
\textbf{Method}
& \textbf{ARPP}
& \textbf{Training AR Factorization}
& \textbf{KV Reuse} \\
\midrule

Teacher Forcing
& \shortstack{ground-truth \\ samples}
& \shortstack{$p(\hat{x}_t^f \mid x^{1:f-1})$}& $\times$ \\
\midrule

\shortstack{Rolling~\cite{liu2025rollingforcing} / Resampling~\cite{guo2025resamplingforcing} /\\Diffusion Forcing~\cite{chen2024diffusionforcing}}
& \shortstack{noisy or resampled \\ ground-truth samples}& \shortstack{$p(\hat{x}_t^f \mid \tilde{x}_{t'}^{1:f-1})$}& $\times$ \\
\midrule

Self Forcing~\cite{huang2025selfforcing}
& \shortstack{self-generated \\ last-step samples}& \shortstack{$p(\hat{x}_t^f \mid \hat{x}_{t'}^{1:f-1})$}& $\checkmark$\\
\midrule

\textbf{Neighbor Forcing(Ours)}
& \shortstack{same-step \\ neighbor reference states}
& \shortstack{$p(\hat{x}_t^f \mid \hat{x}_t^{1:f-1})$}& $\checkmark$ \\

\bottomrule
\end{tabular}
}
\end{table}

This observation suggests that AR diffusion is not inherently incompatible with non-AR backbones. Rather, the critical factor lies in the choice of conditioning representation. Unlike approaches that rely on clean reference frames or artificially noised inputs~\cite{huang2025selfforcing,chen2024diffusionforcing,guo2025resamplingforcing,liu2025rollingforcing}, propagating the latent of a temporally adjacent frame at the same diffusion step introduces a more appropriate inductive bias for AR video generation.
Intuitively, such \textbf{\textit{neighbor latents}} are geometrically close to the current frame on the latent manifold and remain statistically aligned under the same diffusion state. Consequently, they provide temporally informative yet distributionally consistent conditioning signals. In Appendix.~\ref{sec:theory_motivation}, we further provide theoretical justification showing that latents of temporally adjacent frames evaluated at the same diffusion step satisfy a local smoothness property, which makes AR modeling significantly easier and more stable.
Motivated by this insight, we propose \name, a novel AR diffusion formulation built upon \emph{Neighbor Forcing}. Instead of propagating sample-level predictions or heterogeneous noisy histories, \name\ propagates \textbf{\textit{neighbor latents}}—\textit{reference states from temporally adjacent frames evaluated under an identical diffusion step—along the AR chain}. At each denoising step, both the target frame and all preceding reference frames are conditioned under the same noise regime, ensuring that temporal dependencies are modeled within a consistent diffusion state.
This step-consistent design produces a cleaner training signal, improves optimization stability, accelerates convergence, and naturally enables efficient streaming inference via direct reuse of cached key--value representations.

While this step-consistent AR formulation enables streaming generation, scaling to hour-level videos poses a fundamental challenge: the KV cache cannot grow indefinitely. Memory and computation constraints require truncation, preventing access to distant past information and weakening long-term temporal consistency.  Most existing AR methods rely on overlapping windows (e.g., motion-frame overlap), which discard information outside the overlap region and thus progressively forget earlier content. Recent works instead introduce additional memory compression modules~\cite{zhang2025pretraining, zhang2025frame} to summarize historical latents. However, these approaches add architectural complexity and extra encoding steps, limiting their practicality for real-time applications such as human animation ~\cite{xue2024human}.

Meanwhile, prior studies show that pretrained Transformer backbones already exhibit strong capabilities for extracting and organizing temporal features without specialized memory modules~\cite{li2022exploring,cheng2022xmem}. This suggests that long-term context may be efficiently summarized through lightweight mechanisms built upon the multi-layer features of pretrained models.
Inspired by this, and building on Neighbor Forcing, we propose \textbf{ConvKV Memory}, an efficient plug-in memory compression module for AR diffusion. Instead of introducing a complex memory encoder,  ConvKV Memory leverages the pretrained DiT Transformer together with a lightweight 1D-convolution to address long-term memory challenges in hour-scale human animation, such as maintaining consistency of background, clothing and accessories.

Crucially, the local smoothness and step-consistent conditioning induced by Neighbor Forcing make historical key--value representations highly compressible. As a result, ConvKV Memory summarizes past temporal context into a fixed-length memory with minimal information loss. During inference, this compressed memory replaces the growing KV cache, enabling constant-memory AR generation. Importantly, ConvKV Memory introduces only negligible computational overhead, increasing inference time by merely 1.9\%, making it well-suited for streaming and real-time long-video generation.

Overall, this work makes the following contributions.
\textbf{(I)} We identify diffusion-step-aligned neighbor latents as a key inductive bias for AR diffusion, providing a principled and theoretically grounded \textbf{Neighbor Forcing} for step-consistent AR video generation.
\textbf{(II)} We introduce \textbf{ConvKV Memory}, a lightweight plug-in compression mechanism that enables constant-memory hour-scale video generation with negligible overhead.
\textbf{(III)} We develop an optimized real-time system that achieves 20 FPS using only \textbf{two} H100/H200 GPUs with \textbf{end-to-end adaptive FP8 precision, sequence parallelism, and operator fusion} at 720×416 or 512×512 resolution. \name requiring 27.2 TFLOPs per frame—significantly at 512×512 resolution reducing computational cost compared to prior AR diffusion methods. 

\section{Methodology}
\label{sec:method}

\subsection{Overall}
\label{sec:overall}


\begin{figure}[t!]
  \centering
  \includegraphics[width=\linewidth]{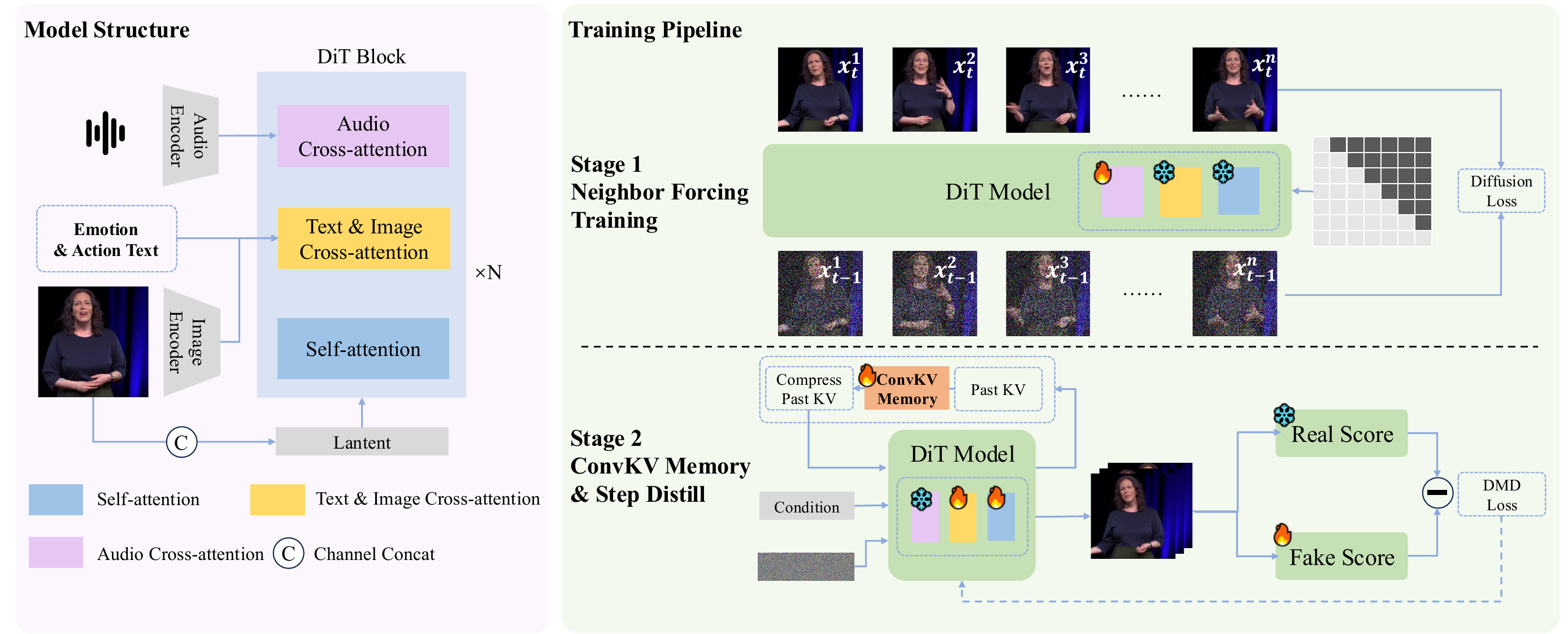}
  \caption{\textbf{Overall training pipeline of \name.}
  \textbf{Left}: The architecture of DiT block.
  \textbf{Right}: The training pipeline consists of two stages: (i) training with step-aligned noisy references and a diffusion loss computed at the same step, and (ii) joint training of ConvKV memory and step distill.
  }
  \label{fig:overall_pipeline}
\end{figure}

\textbf{Model Architecture.}
\label{sec:arch}
\name employs DiT~\citep{peebles2023scalable} with Flow Matching~\citep{liu2022flow, lipman2022flow}, and incorporates audio cross-attention to inject audio conditioning which is shown in Figure~\ref{fig:overall_pipeline}. Frames are encoded into latent chunks  $\{z^1,...z^F\}$ using 3D VAE, and generation is performed in latent space using a denoising network $G_\theta$.

To achieve streaming generation, \name employs a block-wise AR diffusion strategy, in which the sequence $\{z^1,...z^F\}$ is partitioned into blocks of $m$ consecutive chunks, with each block represented by a latent variable $x^n$. Accordingly, the generation process can be formulated as
\begin{align}
\label{5}
\hat{x}^{1:N} =\{\Psi_{T:0}\left(\text{G}_{\theta},\ c^n, \ t \right)| n = 1,..., N\}, \ c^n=\{x_{ref},\ x_t^{1:n-1},\ c_{audio},\quad c_{text}\}
\end{align}
where $\hat{x}^n$ denotes the predicted latents at index $n$; $x_{ref}$ represents the latent of the reference image; $c_{audio}$ and $c_{text}$ denote the audio and text conditioning signals. $\Psi$ denotes the integrator (e.g., UniPC Solver~\citep{zhao2023unipc}) , that simulates the forward diffusion process from $T$-th step to 0-th step to obtain $\hat{x}$.


\textbf{Training and Inference.}
As shown in Figure~\ref{fig:overall_pipeline}, the training pipeline consists of two stages: Neighbor Forcing training and ConvKV Memory\&Step Distill training. In addition, our framework incorporates an auxiliary \textbf{Emotion and Action Editing Module} to enable controllable manipulation of facial expressions and gestures. The first stage adopts the \textbf{Neighbor Forcing} AR formulation to train the alignment between audio and text conditions (e.g., emotion and action prompts) with the generated video, ensuring consistency in lip movements, gestures, and emotional expression. The second stage introduces the \textbf{ConvKV Memory} compression mechanism into the rollout of DMD~\citep{yin2024one} distillation training so that, at inference time, the KV cache remains bounded to a fixed length, enabling stable infinite-length video generation. By integrating the KV cache under the Neighbor Forcing–based autoregressive formulation with the efficient compression mechanism of ConvKV Memory, together with the controllable editing module, SoulX-LiveAct enables real-time hour-scale and even unbounded video generation.




\begin{figure}[t]
  \centering
  \includegraphics[width=\linewidth]{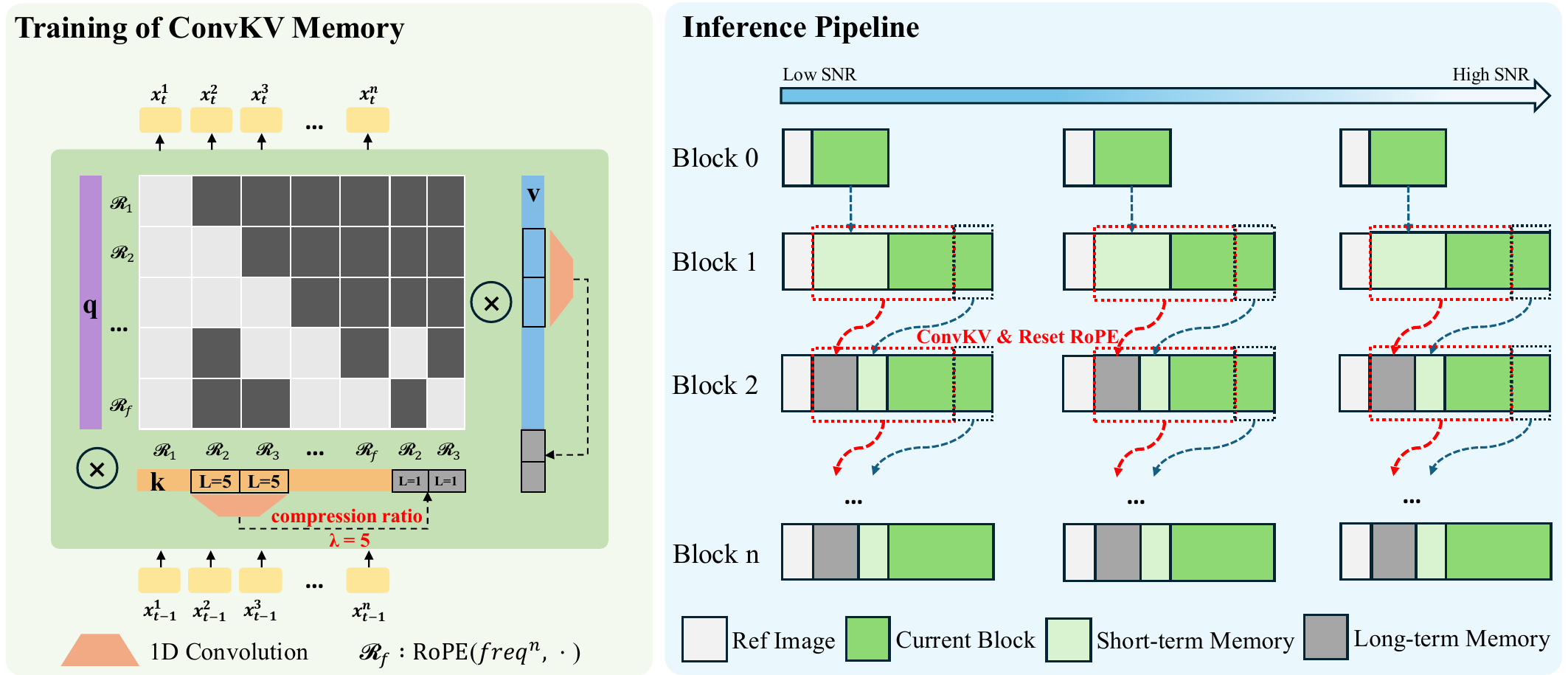}
  \caption{\textbf{Memory mechanism.}
  \textbf{Left}: The KV states corresponding to the long-term memory are compressed via a lightweight 1D convolution operator; 
 \textbf{Right}: The final inference pipeline unifies the Neighbor Forcing formulation and the ConvKV Memory mechanism.}
  \label{fig:memory_arch}
\end{figure}

\subsection{Neighbor Forcing: Step-Aligned AR Propagation}
\label{sec:neighbor_forcing}

A key question in AR diffusion is what to propagate along the chain.
Prior forcing strategies often propagate sample-level states at heterogeneous diffusion steps, requiring the model to align temporal dependencies across mismatched noise semantics.
In contrast, Neighbor Forcing propagates \emph{reference states at the same diffusion step} as the target.

\paragraph{Step-aligned conditioning.}
Let $x_0^{n}$ be the clean latent for block of frame latent $n$.
For a sampled diffusion step $t$, we construct noisy latents using the standard forward process of Flow Matching:
\begin{equation}
x_t^{n} = (1-t)x_0^{n} + t \epsilon^n,
\quad
\epsilon^n \sim \mathcal{N}(0, I).
\label{eq:forward_noising_method}
\end{equation}
At block $n$, Neighbor Forcing  trains the denoiser at step $t$ conditioned on same-step references under the diffusion loss:
\begin{equation}
\mathcal{L}(\theta)
=
\mathbb{E}_{n \sim \text{U}(0, N), t \sim \text{U}(0, 1)}\big[
\|(\epsilon^n - x^n) - G_\theta(x_t^{n}, t \mid x_t^{\,1:n-1})\|^2
\big].
\label{eq:training_loss}
\end{equation}
This design is directly supported by Proposition~2 in Appendix~\ref{sec:theory_motivation},
which shows that at a fixed step $t$, the neighborhood structure between temporal neighbors is preserved in expectation up to an additive step-dependent noise floor.
Therefore, Neighbor Forcing learns temporal dependencies within a single noise space, avoiding cross-step alignment.

\paragraph{Training objective.}
The step-aligned conditioning mechanism enables all blocks to be optimized at a shared diffusion step, thereby avoiding the costly reference-state procedures required by prior forcing strategies (e.g., Self-Forcing). The resulting training objective can be expressed as:
\begin{equation}
\mathcal{L}(\theta)
=
\mathbb{E}_{t \sim \text{U}(0, 1)}\big[
\|(\epsilon - x) - G_\theta(x_t, t, Mask)\|^2
\big].
\label{eq:final_training_loss}
\end{equation}
where, the attention mask $Mask$ of block-wise causal attention with block size $m$ is typically defined as:
\begin{equation} 
Mask_{i,j} = 
\begin{cases} 
1, & \text{if } \left\lfloor \dfrac{j}{m} \right\rfloor \leq \left\lfloor \dfrac{i}{m} \right\rfloor, \\
0, & \text{otherwise.} 
\end{cases} 
\end{equation}
The ablation results on block size $m$ (shown in Table~\ref{tab:ab_blocksize}) suggest that setting the first block size to 6 and the subsequent block sizes to 8 yields the best performance.

\subsection{ConvKV Memory: KV Reuse and Compression}
\label{sec:memory}
Neighbor Forcing makes the conditioning context \emph{step-aligned}.
This has an important systems implication: once a previous frame latents are constructed at step $t$, its representation does not need to be recomputed when predicting subsequent frames at the same step. We exploit this KV cache  mechanism across autoregressive steps.

While KV reuse removes redundant recomputation, a naive KV cache still grows linearly with the number of generated frames, which is prohibitive for hour-scale synthesis. We therefore introduce ConvKV Memory, which (i) keeps a small short-term window of recent KV uncompressed for accuracy, and (ii) continuously compresses older KV into a compact long-term memory using a lightweight 1D convolution operator as shown in Figure~\ref{fig:memory_arch}. This yields bounded memory and stable latency while preserving long-range conditioning.

\paragraph{Training of ConvKV Memory}
We compress the KV states to be summarized using a 1D convolution with compression ratio $\lambda = 5$, which reduces every five chunks of KV states into a single chunk.
The compressed long-term memory at step $t$ is formulated as
\begin{equation}
M_t^{s:e} =  \left(Conv_\theta(k_t^{s:e}), Conv_\theta(v_t^{s:e})\right), \ kernel=stride=\lambda
\end{equation}
where $s$ and $e$ denote the start and end positions of the KV states before compression, and \textit{kernel} and \textit{stride} correspond to the kernel size and stride of the 1D convolution, respectively.

Since position encoding must remain consistent after compression, the compressed keys and values are further processed with a RoPE reset operation, aligning their positional encoding to the starting position $s$. The memory representation therefore becomes
\begin{equation}
M_t^{s:e} =  \left(RoPE(Conv_\theta(k_t^{s:e}), \ frep^s), RoPE(Conv_\theta(v_t^{s:e}), \ frep^s)\right)
\end{equation}
Notably, during training, as the original KV states cannot be altered directly, we append the compressed memory $M_t^{s:e}$ to the end of the key and value tensors. As shown in Figure~\ref{fig:memory_arch} \textbf{Left}, a corresponding modification of the attention mask is applied to properly integrate the ConvKV Memory mechanism.

\paragraph{Inference Pipeline}
During inference, as shown in Figure~\ref{fig:memory_arch} \textbf{Right}, KV compression is not performed for the first two generated latent blocks. Starting from the third iteration onward, the KV states are partitioned into three components: the reference image state, long-term memory,  Short-term Memory and the current block. The long-term memory is defined in Eq. (7).

The ablation results indicate that the KV memory context consists of 6 chunks of frame latents: two chunks represent the reference image, two chunks correspond to long-term memory, and the final two chunks constitute short-term memory by preserving the last two frame-latent chunks from the previous iteration. Owing to the ConvKV Memory mechanism, the KV states remain unchanged within a step-aligned prediction pass, enabling KV reuse to operate in an \emph{append-only} manner: for each newly generated frame, KV is computed only for the appended tokens, while previously cached KV states are reused.



\section{Experiment}
\label{sec:exp}

\noindent \textbf{Implementation Details} \name initializes the self-attention and text\&image cross-attention layers of DiT from Wan2.1~\citep{wan2025wan}, while the audio cross-attention layers is initialized from InfiniteTalk~\citep{yang2025infinitetalk}. The model employs a block-wise AR diffusion framework incorporating Neighbor Forcing and ConvKV Memory.
In Stage 1, we optimize the audio cross-attention module using 300 hours of multimodal paired data composed of video, audio, and emotion/action captions.
In Stage 2, ConvKV Memory is integrated into DMD-style distillation and jointly optimized under a 3-step inference setting for 400 training steps.

\noindent \textbf{Test datasets and evaluation metrics.}
We evaluate our method across diverse scenarios using two evaluation datasets: HDTF~\cite{zhang2021flow}, which focuses on facial dynamics and EMTD~\cite{meng2025echomimicv2}, which incorporates full-body movements. We construct a test set of 100 videos by randomly sampling 50 videos per dataset. We evaluate at 512×512 resolution on HDTF and EMTD. For quantitative evaluation, we adopt Fréchet Inception Distance (FID)~\cite{heusel2017gans} and Fréchet Video Distance (FVD)~\cite{unterthiner2018towards} to assess the \textbf{distributional similarity between generated and real videos}, SyncNet-based Sync-C (confidence) and Sync-D (lip distance)~\cite{prajwal2020lip} to evaluate \textbf{lip synchronization}. We also report the Temporal Quality and Image Quality scores from VBench~\cite{huang2024vbench}as well as the Human Fidelity score from VBench-2.0~\cite{zheng2025vbench} to provide a more comprehensive evaluation of \textbf{overall video quality and human animation quality}.


\begin{table*}[t!]
\centering
\caption{Quantitative comparison on lip-sync accuracy and video quality metrics.}
\label{tab:lip_video_metrics}
\resizebox{\linewidth}{!}{
\begin{tabular}{l l cc cc cc c}
\toprule
\multirow{2}{*}{\textbf{Dataset}} 
& \multirow{2}{*}{\textbf{Model}} 
& \multicolumn{2}{c}{\textbf{Lip-sync Accuracy}} 
& \multicolumn{2}{c}{\textbf{Distribution Similarity}} 
& \multicolumn{2}{c}{\textbf{VBench}} 
& \textbf{VBench-2.0}\\
\cmidrule(lr){3-4} \cmidrule(lr){5-6} \cmidrule(lr){7-8} \cmidrule(lr){9-9}
& 
& Sync-C$\uparrow$ 
& Sync-D$\downarrow$ 
& FID$\downarrow$ 
& FVD$\downarrow$ 
& Temporal Quality$\uparrow$ 
& Image Quality$\uparrow$ 
& Human Fidelity$\uparrow$ \\
\midrule

\multirow{4}{*}{HDTF}
& OmniAvatar~\citep{gan2025omniavatar}                          &  5.13&  10.19&  27.90&  268.47&  86.1&  61.6&  96.8\\
& InfiniteTalk~\citep{yang2025infinitetalk}                     &  7.12&  8.01&  18.15&  169.88&  94.5&  59.9&  99.4\\
& Live-Avatar~\citep{huang2025liveavatarstreamingrealtime}      &  7.68&  8.38&  15.85&  206.20&  91.8&  59.2&  99.8\\
& Ours  &  \textbf{9.40}&  \textbf{6.76}&  \textbf{10.05}&  \textbf{69.43}&  \textbf{97.6}&  \textbf{63.0}&  \textbf{99.9}\\
\midrule

\multirow{4}{*}{EMTD}
& OmniAvatar~\citep{gan2025omniavatar}                          &  6.24&  8.63&  \textbf{33.63}&  \textbf{589.5}&  91.9&  63.2&  96.7\\
& InfiniteTalk~\citep{yang2025infinitetalk}                     &  7.98&  7.44&  37.26&  339.0&  94.9&  61.6&  96.7\\
& Live-Avatar~\citep{huang2025liveavatarstreamingrealtime}      &  6.93&  8.23&  42.52&  365.0&  93.6&  63.0&  96.6\\
& Ours  &  \textbf{8.61}&  \textbf{7.29}&  80.90&  771.6&  \textbf{97.3}&  \textbf{65.7}&  \textbf{98.9}\\
\bottomrule
\end{tabular}
}
\end{table*}

\subsection{Quantitative Experiments}


\paragraph{HDTF~\cite{zhang2021flow}.}
As a portrait-level benchmark, HDTF mainly assesses lip-sync precision and visual realism. \name achieves substantial improvements in Sync-C, Sync-D, FID, and FVD, demonstrating superior synchronization accuracy and generation quality. Furthermore, the strong performance on both VBench and VBench-2.0 indicates that the model effectively preserves identity and structural consistency while maintaining stable motion dynamics.

\paragraph{EMTD~\cite{meng2025echomimicv2}.}
\name achieves competitive results across all VBench metrics and attains 97.6 in Temporal Quality and 63.0 in Image Quality, outperforming previous methods. Under VBench-2.0, the model obtains 96.6 on Human\_Anatomy and perfect scores (1.0) on both Human\_Clothes and Human\_Identity, highlighting strong identity preservation and fine-grained animation quality. In addition, it achieves the best lip-sync accuracy, further demonstrating its robustness and generalization capability across datasets.

\paragraph{Efficiency comparison.}
Beyond generation quality, we further compare computational efficiency among different methods. We report throughput (FPS), the number of GPUs required for real-time inference, and computational cost measured by TFLOPs per frame. In terms of computational complexity, our approach requires 27.2 TFLOPs per 512$\times$512 frame, which is significantly lower than Bidirectional baselines (50.2 TFLOPs/frame) and AR counterpart Live-Avatar (39.1 TFLOPs/frame). Owing to the adoption of end-to-end FP8 precision, our method reaches 20 videos per second using only two H100/H200 GPUs, achieving real-time performance at 512×512 or 720×416 resolution with high hardware efficiency.
These results demonstrate that our model not only achieves strong visual fidelity, but also maintains favorable efficiency-performance trade-offs suitable for real-time deployment.


\renewcommand{\thefootnote}{\dag}
\begin{table}[ht!]
    \vspace{-5pt}
    \caption{\textbf{Inference efficiency comparison between other real-time methods.}}
    \vspace{+5pt}
    \label{tab:Efficiency_comparison}
    \centering
    \resizebox{0.7\columnwidth}{!}{
    \begin{tabular}{c c c cc}
        \toprule
        \textbf{Model} & \textbf{Thoughput(FPS)} & \textbf{Latency(s) }$\downarrow$& \textbf{GPUs }$\downarrow$&\textbf{TFLOPs per frame}$\downarrow$\\
        \midrule
 InfiniteTalk\footnotemark& 25 FPS& 3.20 s&8&50.2\\
 Live-Avatar& 20 FPS& 2.89 s&5&39.1\\
 Neighbor-forcing& 20 FPS& \textbf{0.94 s}&\textbf{2}&\textbf{27.2}\\ \bottomrule
    \end{tabular}
    }
    \vspace{-5pt}
\end{table}
\footnotetext{Applied the LoRA-weighted 4-step distillation provided by LightX2V.}

\subsection{Qualitative Results}
\begin{figure*}
     \centering
     \includegraphics[width=1.0\linewidth]{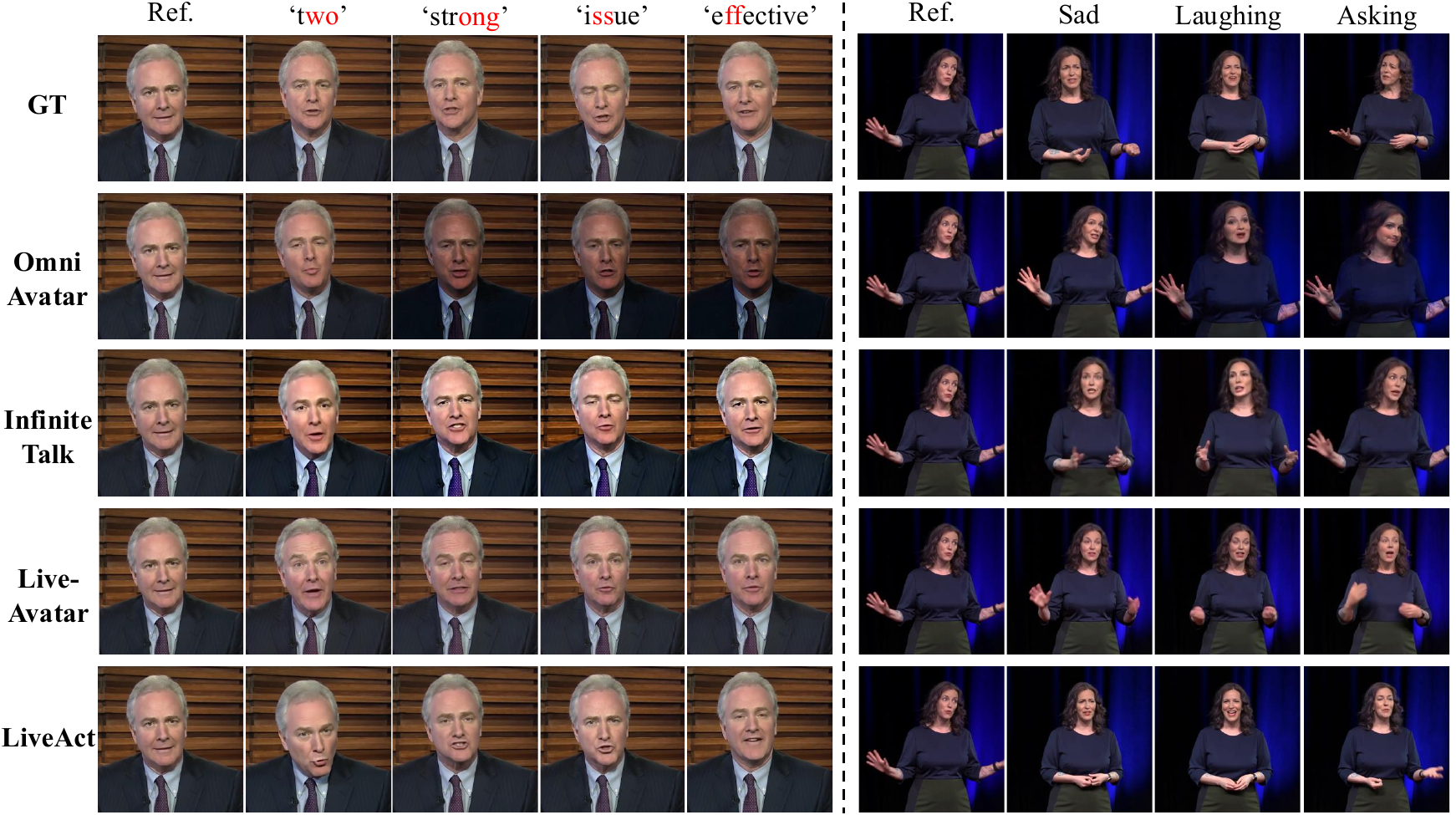}
    \caption{Qualitative comparison of \textbf{\textit{lip-motion accuracy and emotion–action coordination}}. 
\name achieves more precise lip–phoneme alignment and maintains coherent facial expressions and body movements under emotion–action interactions, while baseline methods show misalignment or temporal jitter.}
    \label{fig:lip_action_quality} 
\end{figure*}

\paragraph{Lip-motion and emotion-action coordination comparison.}
We also provide fine-grained comparisons on lip motion and emotion-action synchronization. As shown in Figure~\ref{fig:lip_action_quality}, \name produces more precise mouth shapes that better match the articulated words, especially for bilabial and open-vowel phonemes. Furthermore, in challenging scenarios involving emotion-action interactions (e.g., partial occlusion by gestures), it maintains facial expressions and body movements that closely resemble those in real videos. In contrast, baseline methods often exhibit lip–phoneme misalignment or temporal jitter.

\begin{figure*}
     \centering
     \includegraphics[width=1.0\linewidth]{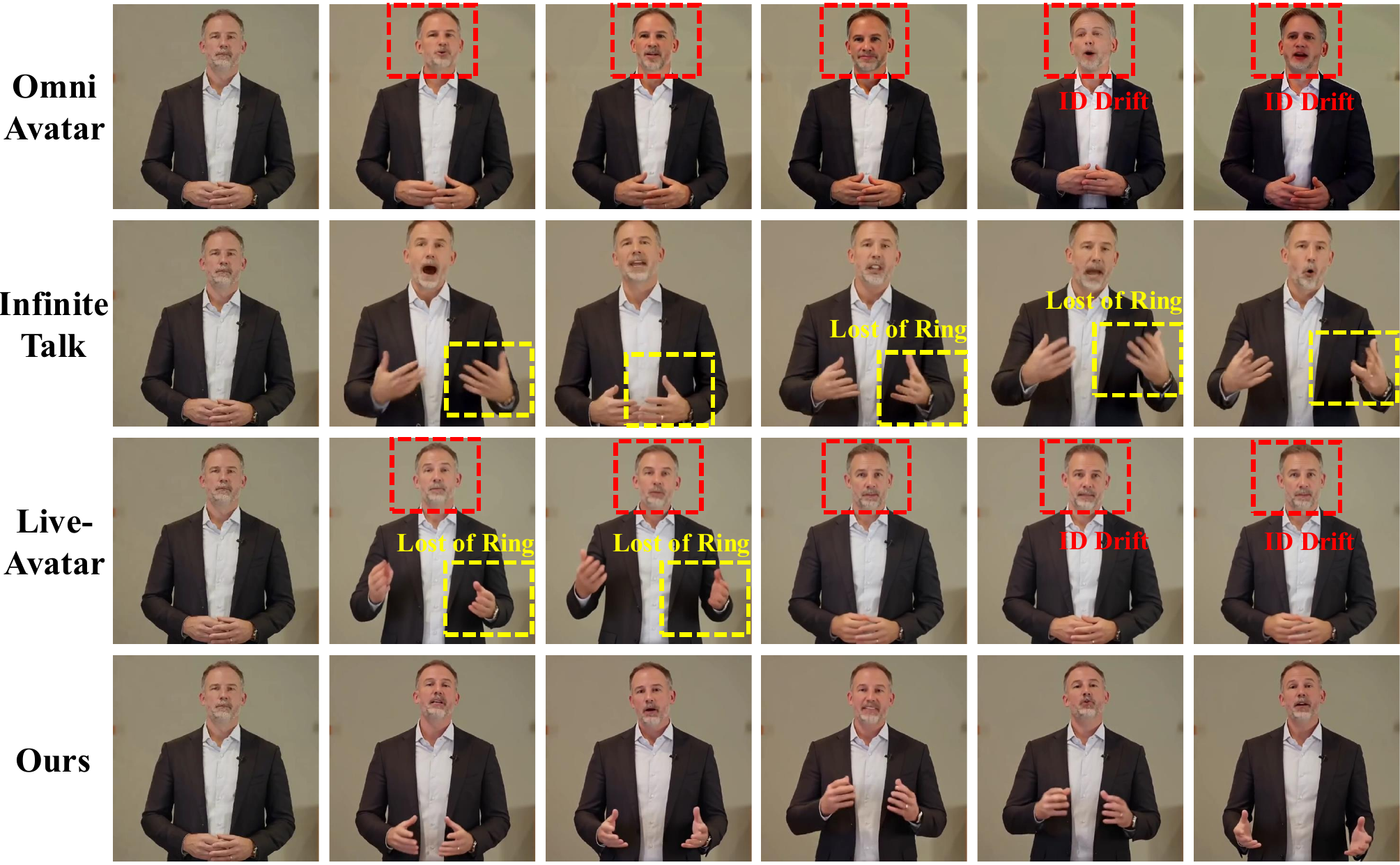}
    \caption{\textbf{\textit{Long-video consistency}} comparison. 
Red boxes indicate identity drift (notably severe in \textit{OmniAvatar}), while yellow boxes highlight fine-grained detail inconsistencies (e.g., missing accessories in \textit{InfiniteTalk} and \textit{Live-Avatar}). 
\name maintains stable identity and details throughout the sequence.}
    \label{fig:longvid} 
\end{figure*}

\paragraph{Long-video consistency analysis.}
To evaluate long-term temporal stability, we conduct a qualitative comparison on long-video generation, focusing on identity consistency and fine-grained detail preservation, as shown in Figure~\ref{fig:longvid}. As highlighted by the red dashed boxes, both OmniAvatar and Live-Avatar exhibit noticeable identity drift over time. In particular, OmniAvatar suffers from accumulated errors as the sequence progresses, resulting in gradual deviation in facial structure and appearance. This phenomenon indicates limited long-term temporal stability in identity preservation. As indicated by the yellow dashed boxes, InfiniteTalk and Live-Avatar show inconsistencies in person-related fine-grained details. For example, accessories such as rings intermittently disappear or reappear across frames. Such instability reflects insufficient constraints on identity-related attributes during long-form generation. In contrast, our method maintains stable identity representation and preserves fine-grained details consistently throughout the entire sequence, demonstrating superior long-term temporal coherence.

Overall, the qualitative results complement the quantitative findings, demonstrating that our approach achieves strong long-term consistency and accurate lip–speech synchronization in realistic scenarios.

\subsection{Ablation Studies}

\begin{figure*}
     \centering
     \includegraphics[width=1.0\linewidth]{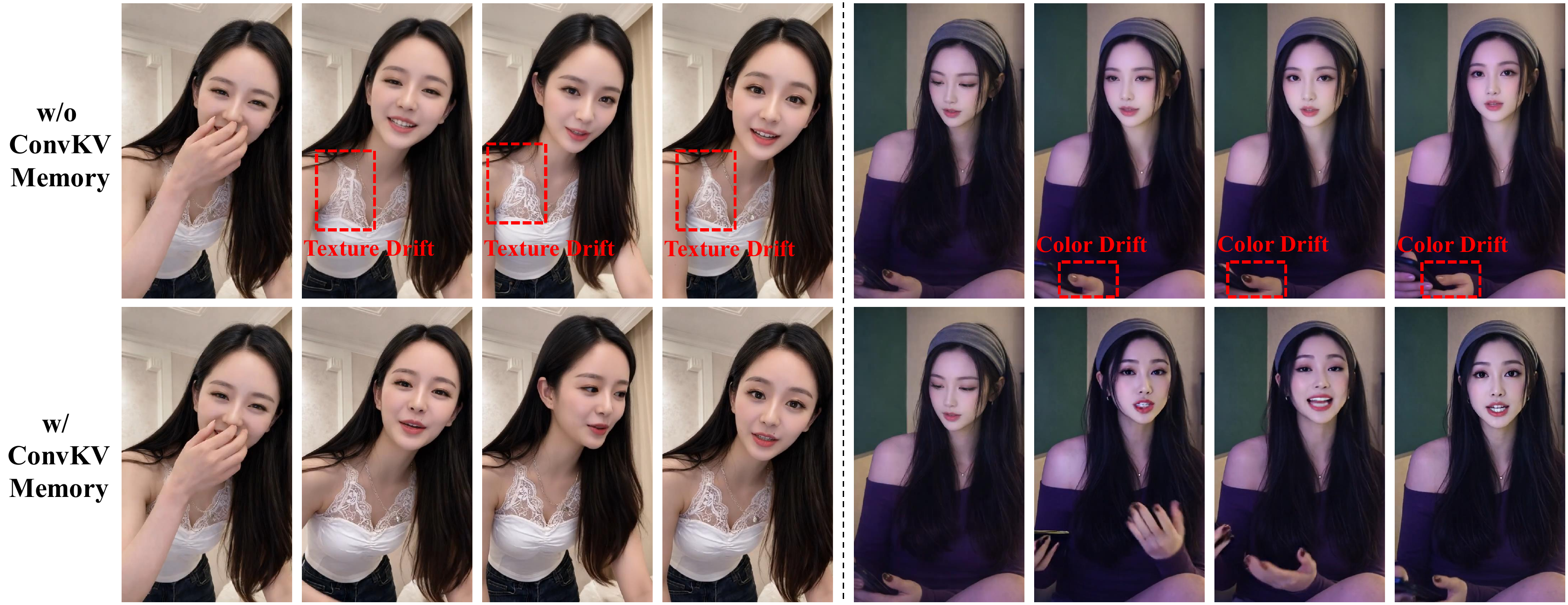}
    \caption{\textbf{\textit{ConvKV Memory Mechanism consistency}} comparison. 
Red boxes highlight inconsistencies in clothing or hand details (e.g., inconsistencies in clothing patterns and nail color).
The ConvKV Memory mechanism ensures stable identity and detail preservation throughout the sequence.}
     \vspace{-12pt}
    \label{fig:ab_memory} 
\end{figure*}

\paragraph{Effect of ConvKV Memory Mechanism.}
We further analyze the role of the ConvKV memory mechanism in preserving long-range consistency as shown in Figure~\ref{fig:ab_memory}. Qualitative comparisons without the ConvKV Memory mechanism exhibit inconsistencies in clothing or hand details, demonstrating that the ConvKV Memory mechanism ensures stable identity and detail preservation throughout the sequence.

\begin{figure*}
     \centering
     \includegraphics[width=0.8\linewidth]{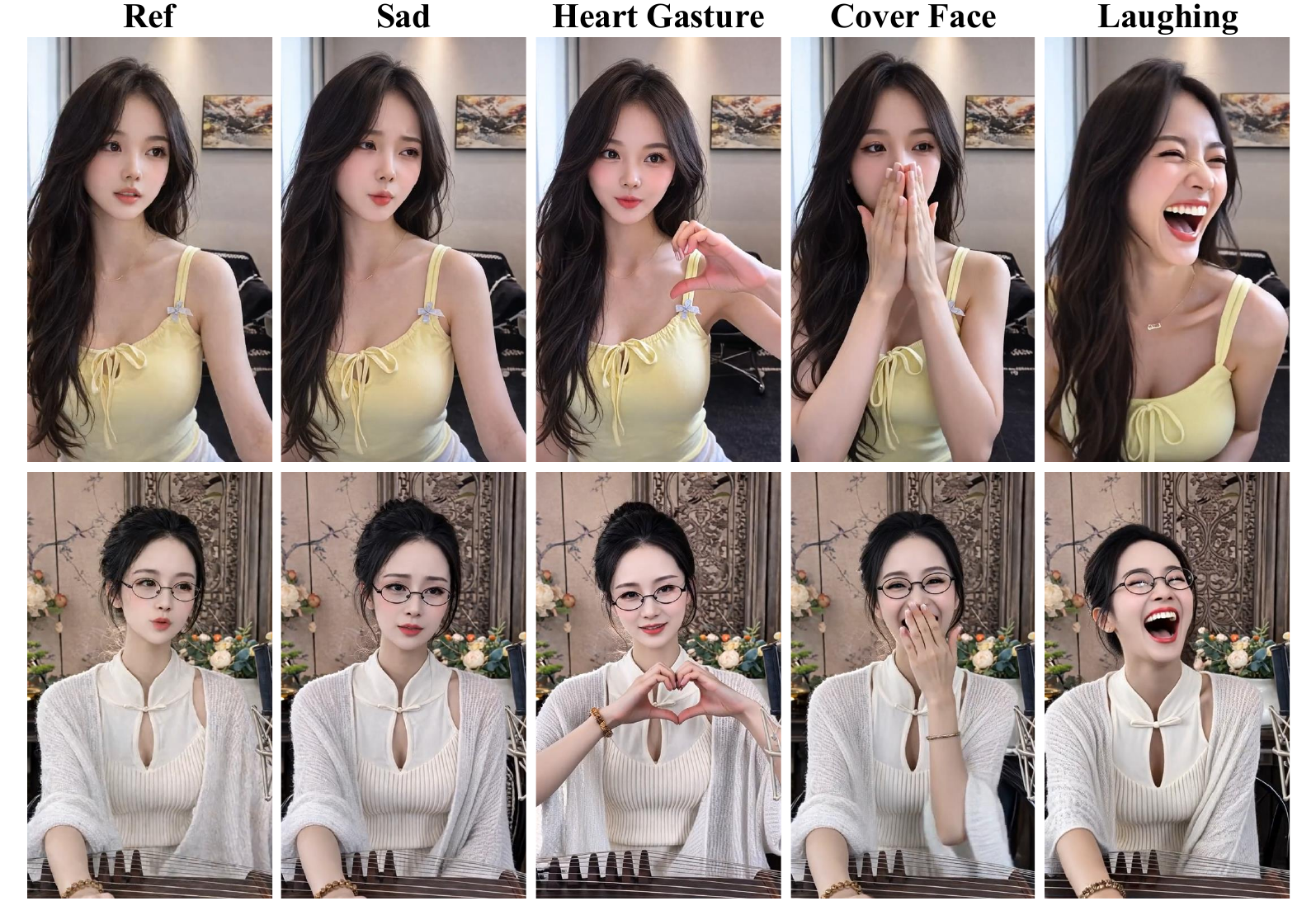}
\caption{\textbf{\textit{Effect of the Emotion and Action Editing Module}}. 
Given a reference image (Ref), our motion editing module enables controllable modification of facial expressions and gestures, including sad expression, heart gesture, covering face, and laughing. 
}
    \label{fig:text_editing} 
\end{figure*}

\paragraph{Effect of Emotion and Action Editing Module.}
To evaluate the motion editing capability, we provide visual comparisons in Figure~\ref{fig:text_editing}. We display representative frames arranged horizontally to illustrate the editing process over time. Our motion editing module enables controllable modification of head pose and gesture dynamics while preserving identity and lip synchronization. Compared to the baseline without motion editing, our method produces smoother transitions and avoids unnatural deformation or motion artifacts. These results demonstrate that the motion editing module provides flexible yet stable motion control.




\paragraph{Training cost difference between Neighbor Forcing and Self Forcing.}
Due to the AR paradigm of Self Forcing, significant modifications are required for bidirectional models. As a result, it necessitates a large amount of training data (e.g., 16K) for 'ODE initialization' training and 1000 steps of distillation. In contrast, Neighbor Forcing propagates the latent of a temporally adjacent frame at the same diffusion step, eliminating the need for complex ODE initialization training and requiring only 300 steps of distillation, as in Table~\ref{tab:diff_selfforcing_ours}.

\begin{table}[h!]
    \centering    
    \vspace{+5pt}
    \caption{\textbf{Difference between Neighbor Forcing and Self Forcing.}}
    \label{tab:diff_selfforcing_ours}
\begin{tabular}{ccc}
\hline
\textbf{Method}  & \textbf{ODE Init. Training} & \textbf{DMD Distill Training} \\ \hline
Self Forcing     & Yes                         & 1000 Steps                    \\ \hline
Neighbor Forcing & No                          & 300 Steps                     \\ \hline
\end{tabular}
\end{table}

\paragraph{Ablation study on the memory-block size and current-block size hyperparameters.}
To meet the real-time constraint of 20 FPS throughput, the model is required to maintain an average per-frame generation latency below 50 ms. Accordingly, we perform an ablation study on the memory-block and current-block size hyperparameters.

\begin{table}[h!]
    \centering    
    \vspace{+5pt}
    \caption{\textbf{Ablation study on the memory-block size and current-block size hyperparameters.}}
    \label{tab:ab_blocksize}
\begin{tabular}{cccc}
\hline
\textbf{Memory-block Size} & \textbf{Current-block Size} & \textbf{Cost/frame} & \textbf{Meat Real-time Requirement} \\ \hline
6                          & 6                           & 52ms                & No                         \\ \hline
8                          & 8                           & 60ms                & No                         \\ \hline
6                          & 8                           & 49ms                & Yes                        \\ \hline
\end{tabular}
\end{table}

Overall, the ablation experiments confirm that each component contributes to either training stability, temporal consistency, or controllable motion generation, and their combination leads to the best overall performance.
\section{Conclusion} 
In this work, we revisit AR diffusion from the perspective of representation propagation along the AR chain. We show that diffusion-step-aligned neighbor latents provide a principled and effective inductive bias for AR video generation, enabling stable optimization and temporally coherent streaming inference even when built upon pretrained non-AR backbones. Based on this insight, we propose \name, a step-consistent AR diffusion formulation, together with ConvKV Memory, a lightweight plug-in compression module that enables constant-memory long-video generation with negligible computational overhead. Extensive experiments demonstrate strong lip-sync accuracy, visual fidelity, long-term identity consistency, and real-time efficiency.


\bibliographystyle{abbrv} 
\bibliography{references}  

\clearpage
\appendix
\textbf{\large\bfseries Appendix}
\section{Local Smoothness and Step-Consistent Neighbors}
\label{sec:theory_motivation}

\paragraph{Setup.}
Let $\{x^{f}\}_{f=1}^{F}$ be a video sequence in $\mathbb{R}^D$, 
and let $z_0^{\,f} = E(x^{f}) \in \mathbb{R}^d$ denote its clean latent representation 
through an encoder $E$.

We reserve subscript $t$ for diffusion steps and superscript $f$ for frame indices.
For diffusion step $t$, define the forward noising process
\begin{equation}
z_t^{\,f} = \alpha_t z_0^{\,f} + \sigma_t \epsilon^{\,f},
\qquad 
\epsilon^{\,f} \sim \mathcal{N}(0,I_d),
\label{eq:forward_noising}
\end{equation}
where $(\alpha_t,\sigma_t)$ follow a standard noise schedule.

\paragraph{Assumptions (manifold and regularity).}

\begin{assumption}[Low-dimensional state with bounded rendering error]
\label{assump:render_error}
There exist latent states $u^{f} \in \mathbb{R}^m$ with $m \ll D$ 
and a rendering map $g:\mathbb{R}^m \rightarrow \mathbb{R}^D$ such that
\begin{equation}
x^{f} = g(u^{f}) + r^{f}, 
\qquad 
\|r^{f}\| \le \varepsilon_r.
\label{eq:render_error}
\end{equation}
Moreover, the dynamics is locally smooth with bounded velocity:
\begin{equation}
\|u^{f+1}-u^{f}\| \le \Delta_u 
\quad \text{for most } f.
\label{eq:bounded_velocity}
\end{equation}
\end{assumption}

\begin{assumption}[Local Lipschitzness near the data manifold]
\label{assump:lipschitz}
There exist constants $L_E, L_g > 0$ such that:
\begin{equation}
\|E(x)-E(x')\| \le L_E \|x-x'\|
\quad \text{for all } x,x' \in \mathcal{N}_{\rho}(\mathcal{M}),
\label{eq:lipschitz_E}
\end{equation}
and
\begin{equation}
\|g(u)-g(u')\| \le L_g \|u-u'\|
\quad \text{for all } u,u' \in \mathcal{U},
\label{eq:lipschitz_g}
\end{equation}
where $\mathcal{M}=\{g(u):u\in\mathcal{U}\}$ is the data manifold and 
$\mathcal{N}_{\rho}(\mathcal{M})$ denotes its $\rho$-neighborhood containing $\{x^{f}\}$.
\end{assumption}

\paragraph{Proposition 1 (Temporal neighbors are latent neighbors).}
Under Assumptions~\ref{assump:render_error}--\ref{assump:lipschitz}, 
for any $f$ such that Eq.~\eqref{eq:bounded_velocity} holds and 
$x^{f},x^{f+1}\in\mathcal{N}_{\rho}(\mathcal{M})$,
we have
\begin{equation}
\|z_0^{\,f+1}-z_0^{\,f}\|
\le L_E\big(L_g\Delta_u + 2\varepsilon_r\big).
\label{eq:latent_neighbor_bound}
\end{equation}

\noindent\textit{Proof.}
By Lipschitzness of $E$ on $\mathcal{N}_{\rho}(\mathcal{M})$,
\begin{equation}
\|z_0^{\,f+1}-z_0^{\,f}\|
= \|E(x^{f+1})-E(x^{f})\|
\le L_E\|x^{f+1}-x^{f}\|.
\end{equation}
Using the decomposition $x^{f}=g(u^{f})+r^{f}$,
\begin{align}
\|x^{f+1}-x^{f}\|
&= \|g(u^{f+1})-g(u^{f}) + (r^{f+1}-r^{f})\| \\
&\le \|g(u^{f+1})-g(u^{f})\| + \|r^{f+1}-r^{f}\| \\
&\le L_g\|u^{f+1}-u^{f}\| + \|r^{f+1}\|+\|r^{f}\| \\
&\le L_g\Delta_u + 2\varepsilon_r,
\end{align}
where we used Eqs.~\eqref{eq:lipschitz_g}--\eqref{eq:bounded_velocity}.
Combining the inequalities yields Eq.~\eqref{eq:latent_neighbor_bound}.
\hfill$\square$

\paragraph{Proposition 2 (Same-step noising preserves step-consistent neighborhood in expectation).}
Assume $\epsilon^{\,f}$ are i.i.d.\ $\mathcal{N}(0,I_d)$ and independent of $\{z_0^{\,f}\}$.
Then for any fixed diffusion step $t$,
\begin{equation}
\mathbb{E}\,\|z_t^{\,f+1} - z_t^{\,f}\|^2
= \alpha_t^2 \|z_0^{\,f+1}-z_0^{\,f}\|^2 + 2\sigma_t^2 d.
\label{eq:same_step_distance}
\end{equation}

\noindent\textit{Proof.}
From Eq.~\eqref{eq:forward_noising},
\begin{equation}
z_t^{\,f+1} - z_t^{\,f}
= \alpha_t \Delta z_0^{\,f} + \sigma_t \Delta \epsilon^{\,f},
\end{equation}
where 
\(
\Delta z_0^{\,f} := z_0^{\,f+1}-z_0^{\,f}
\)
and
\(
\Delta \epsilon^{\,f} := \epsilon^{\,f+1}-\epsilon^{\,f}.
\)

Expanding the squared norm,
\begin{equation}
\|\alpha_t\Delta z_0^{\,f} + \sigma_t\Delta\epsilon^{\,f}\|^2
= \alpha_t^2\|\Delta z_0^{\,f}\|^2
+ \sigma_t^2\|\Delta\epsilon^{\,f}\|^2
+ 2\alpha_t\sigma_t
\langle \Delta z_0^{\,f}, \Delta\epsilon^{\,f}\rangle.
\end{equation}

Taking expectation and using independence together with
$\mathbb{E}[\Delta\epsilon^{\,f}]=0$, we get
\[
\mathbb{E}\langle \Delta z_0^{\,f}, \Delta\epsilon^{\,f}\rangle = 0.
\]
Since $\epsilon^{\,f+1},\epsilon^{\,f}$ are i.i.d.\ $\mathcal{N}(0,I_d)$,
\begin{align}
\mathbb{E}\|\Delta\epsilon^{\,f}\|^2
&= \mathbb{E}\|\epsilon^{\,f+1}-\epsilon^{\,f}\|^2 \\
&= \mathbb{E}\|\epsilon^{\,f+1}\|^2
 + \mathbb{E}\|\epsilon^{\,f}\|^2
 - 2\mathbb{E}\langle\epsilon^{\,f+1},\epsilon^{\,f}\rangle \\
&= d + d - 0 = 2d.
\end{align}

Substituting back yields Eq.~\eqref{eq:same_step_distance}.
\hfill$\square$

\paragraph{Implications.}
Proposition~1 shows that under locally smooth dynamics and regular encoders, 
temporally adjacent frames remain close in latent space, up to bounded modeling error.
Proposition~2 shows that at a fixed diffusion step $t$, the expected squared distance 
between noisy temporal neighbors decomposes into a signal term scaled by $\alpha_t^2$ 
and an additive isotropic noise floor $2\sigma_t^2 d$ independent of frame index $f$.
This step-consistent decomposition motivates conditioning on same-step temporal neighbors 
for autoregressive propagation and suggests that short temporal windows admit 
redundancy-aware compression in KV memory without significantly distorting the conditioning signal.

\section{Related Work}
\textbf{Autoregressive Video Generation.} Early methods predominantly employ the Teacher Forcing~\citep{2411.16375, ACDiT, Test-Time} strategy to generate video sequences autoregressively. However, this approach suffers from significant train-test mismatch, leading to accumulated errors during video generation. Some studies~\citep{zhang2025frame, VSA} adopt planning generation strategy, pre-determining distant future frames and then interpolating intermediate ones to mitigate error accumulation inherent in autoregressive processes. Subsequent research introduced Diffusion Forcing~\citep{chen2024diffusionforcing, yin2025causvid, chen2025skyreelsv2infinitelengthfilmgenerative}, which mitigates this mismatch by assigning independent noise levels to sequential frames. More recently, Self Forcing~\citep{huang2025selfforcing} has addressed error accumulation by directly conditioning the model on its own generated historical frames during training. Building upon this, Self-Forcing++~\citep{cui2025self} incorporates Distribution Matching Distillation (DMD) to surpass the temporal constraints of base models. Self-Resampling~\citep{guo2025resamplingforcing} utilizes online-updated weights to autoregressively resample past frames, simulating and counteracting potential model biases during inference to significantly enhance the stability of long-sequence generation. Despite these advancements in mitigating error accumulation, the aforementioned methods often incur substantial computational overhead due to complex sampling strategies or iterative feedback loops, resulting in high training costs. In contrast, we propose Neighbor Forcing, which reuses latents from the same step to significantly improve training efficiency and achieve faster convergence while maintaining superior generation quality.

\textbf{Memory Compression.} Ensuring interframe consistency is paramount in long horizon video generation. Several approaches have introduced history compression mechanisms to balance generation quality and resource consumption. FramePack~\citep{zhang2025frame} encodes preceding frames into fixed size latent space features to serve as context. Lvmin Zhang et al.~\citep{zhang2025pretraining} employed a UNet-based architecture to highly condense redundant historical frame information, enhancing the coherence of extended video sequences.
However, such complex network architectures are not suitable for real-time human animation models. To address this, we propose the ConvKV Memory mechanism, which employs a lightweight 1D convolution on the keys and values in causal attention to manage long-term memory.

\end{document}